\documentclass[10pt,twocolumn,letterpaper]{article}

\usepackage[pagenumbers]{cvpr} 

\usepackage{stfloats}
\usepackage{multirow}
\usepackage{longtable}
\usepackage{appendix}
\usepackage{multicol}

\usepackage{soul}
\definecolor{darkgreen}{RGB}{84, 130, 53}
\definecolor{darkblue}{RGB}{46, 117, 182}
\definecolor{darkred}{RGB}{192, 0, 0}
\definecolor{myhighlightcolor_gray}{RGB}{229,229,229}
\definecolor{myhighlightcolor_brown}{RGB}{245,237,230}

\usepackage{bm}
\usepackage{amsmath}
\usepackage{tcolorbox}
\usepackage{makecell}

\usepackage{tabularray}

\definecolor{cvprblue}{rgb}{0.21,0.49,0.74}
\usepackage[pagebackref,breaklinks,colorlinks,allcolors=cvprblue]{hyperref}

\def\paperID{11236} 
\def\confName{CVPR}
\def\confYear{2025}

\title{MG-MotionLLM: 
A Unified Framework for Motion Comprehension and Generation across Multiple Granularities}

\author{
Bizhu Wu$^{1,2,3}$\quad Jinheng Xie$^{4}$\quad Keming Shen$^{1,3}$\quad Zhe Kong$^5$\quad Jianfeng Ren$^{2}$\thanks{Corresponding authors}\quad Ruibin Bai$^{2}$\quad \\ 
Rong Qu$^6$\quad Linlin Shen$^{1,2,3*}$ \\
$^1$ Computer Vision Institute, School of Computer Science \& Software Engineering, Shenzhen University \\ 
$^2$ School of Computer Science, University of Nottingham Ningbo China, Ningbo, China \\ 
$^3$ Guangdong Provincial Key Laboratory of Intelligent Information Processing \\
$^4$ National University of Singapore \quad $^5$ Shenzhen Campus of Sun Yat-sen University \\ 
$^6$ School of Computer Science, University of Nottingham, Nottingham, United Kingdom \\
{\tt\small wubizhu@email.szu.edu.cn, jianfeng.ren@nottingham.edu.cn, llshen@szu.edu.cn} \\ 
}

\begin{document}
\maketitle


\begin{abstract}
Recent motion-aware large language models have demonstrated promising potential in unifying motion comprehension and generation. 
However, existing approaches primarily focus on coarse-grained motion-text modeling, where text describes the overall semantics of an entire motion sequence in just a few words. 
This limits their ability to handle fine-grained motion-relevant tasks, such as understanding and controlling the movements of specific body parts.
To overcome this limitation, we pioneer MG-MotionLLM, a unified motion-language model for multi-granular motion comprehension and generation.
We further introduce a comprehensive multi-granularity training scheme by incorporating a set of novel auxiliary tasks, such as localizing temporal boundaries of motion segments via detailed text as well as motion detailed captioning, to facilitate mutual reinforcement for motion-text modeling across various levels of granularity. 
Extensive experiments show that our MG-MotionLLM achieves superior performance on classical text-to-motion and motion-to-text tasks, 
and exhibits potential in novel fine-grained motion comprehension and editing tasks. 
Project page: \href{https://github.com/CVI-SZU/MG-MotionLLM}{CVI-SZU/MG-MotionLLM}  
\end{abstract}


\section{Introduction}
\label{sec:intro}

\begin{figure}[t]
    \centering
    \includegraphics[width=1.0\linewidth]{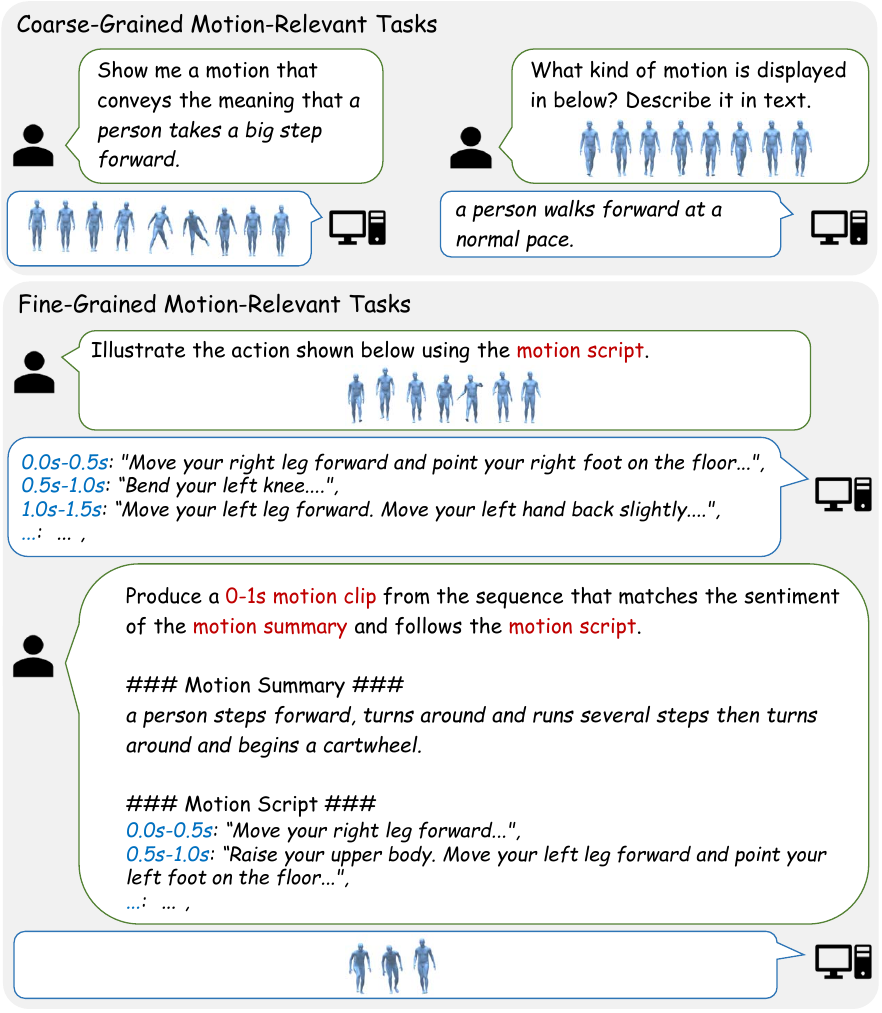}
    \setlength{\abovecaptionskip}{-0.3em}  
    \caption{
        MG-MotionLLM can address diverse motion-relevant tasks at multiple granularities by giving different instructions in a unified manner. 
        We show results for some existing coarse-grained tasks, such as text-to-motion and motion captioning (\textit{upper} block), 
        and newly developed fine-grained tasks, including motion-to-detailed text and motion localization (\textit{bottom} block). 
        The temporal progression of motion is illustrated from left to right. 
        \textcolor{darkgreen}{Green boxes} denote the input, and \textcolor{darkblue}{blue boxes} are the output.
    }
    \label{fig:intro}
\vspace{-1em}
\end{figure}

Motion comprehension and generation are crucial for diverse applications, including AR/VR creation, video games, and virtual reality. 
Numerous studies~\cite{guo2022tm2t, humanml3d, t2mgpt} have been devoted to specific tasks, such as captioning 3D human motions~\cite{guo2022tm2t} and generating 3D human motions from textual descriptions~\cite{mdm, t2mgpt, humanml3d, guo2024momask}, to promote the development of these areas. 
However, the comprehension and generation of motions are often studied separately. 
Recent works~\cite{openmotionlab_motiongpt, wu2024motionllm} have resorted to large language models (LLMs) to construct unified motion-language models that can both generate plausible motions and vivid textual descriptions.

Existing motion-related LLMs~\cite{wu2024motionllm, zhang2024motiongpt, openmotionlab_motiongpt, luo2024m3gpt}, however, often align motions with coarse-grained textual descriptions (\textit{upper} block in Fig.~\ref{fig:intro}), \textbf{ignoring the intricate nuances of motions and precise temporal information}.
Some approaches~\cite{kalakonda2023actiongpt, yazdian2023motionscript, he2023semanticboost, zhang2024finemogen} have attempted to incorporate more detailed descriptions to enhance motion generation.
For example, SemanticBoost~\cite{he2023semanticboost} and MotionScript~\cite{yazdian2023motionscript} focused on the movement of each body part and translated them into predefined statuses to enhance motion generation. 
Action-GPT~\cite{kalakonda2023actiongpt} prompted LLMs to produce detailed descriptions for given action labels.
Despite these advances, such approaches focus solely on detailed motion generations, \textbf{failing to integrate with detailed motion understanding}.

Therefore, in this paper, we explore both fine-grained motion generation and understanding, in which textual descriptions detail body part movements (BPM) over time, referred to as the \textit{motion script}.
And we propose a unified motion-language model, \textbf{MG-MotionLLM}, that leverages the strong language generation capabilities of LLMs to address motion-relevant tasks across \underline{M}ultiple \underline{G}ranularites, as displayed in Fig.~\ref{fig:intro}. 
Specifically, to enable MG-MotionLLM to comprehend and generate human-like motions, we first map motions into discrete tokens via motion VQ-VAE.
We then fine-tune the language model with motion data and texts at various granularities, to leverage strong textual priors in the LLM~\cite{t5},
incorporate motion information, 
and explore the correlation between two modalities.

However, after initial attempts, we found that it is \textbf{challenging to effectively establish connections between motions and long, detailed textual descriptions}.
Specifically, when we directly instruction-tuned the LLM to generate motions from both coarse and detailed descriptions, the model achieved a Top-3 retrieval accuracy of 75.0\%, which was lower than the 77.3\% achieved with only coarse descriptions.
We attribute this performance gap to the massive information present in the detailed descriptions, which may distract the model from capturing the global semantics of the motions.
To address this, we propose a two-stage,
\textbf{comprehensive multi-granularity training scheme}.
In the first stage, termed Granularity-Synergy Pre-training, we introduce coarse-grained motion-relevant tasks to assist fine-grained ones in capturing semantic meanings. 
Additionally, we incorporate a bunch of newly designed auxiliary tasks, such as localizing temporal boundaries of motion segments based on their detailed descriptions and the coarse captions for entire motion sequences. 
These tasks, in turn, enhance the model's understanding of motion details and promote more effective learning of coarse-grained tasks.
In the second stage, we further fine-tune the model to enhance its performance on specific tasks.
Extensive experiments show that MG-MotionLLM excels at both motion understanding and generation across various granularities of text.

In summary, our contributions are as follows:
\textbf{First}, we pioneer a research work on exploring both fine-grained motion generation and understanding, and propose MG-MotionLLM to solve a set of motion-relevant tasks across multiple levels of granularity in a unified framework. 
\textbf{Secondly}, the proposed multi-granularity training scheme with novel auxiliary tasks captures motion-related features at different levels, improving understanding across a wide range of tasks. As shown in experiments, these motion-relevant tasks at different granularities mutually enhance each other's performance, and drive MG-MotionLLM to achieve competitive results on classical coarse-grained motion-relevant tasks.
\textbf{Thirdly}, qualitative results show that our model enables a series of novel fine-grained motion-relevant applications, such as fine-grained motion editing - all achieved within a single model.


\section{Related Work}
\label{sec:related_work}

\subsection{Text-Driven Human Motion Generation}
\label{sec:related_work_t2m}
Significant progress has been made in text-driven motion generation~\cite{t2mgpt, mdm, petrovich2022temos, kim2023flame, barquero2024flowmdm, huang2024como, petrovich24stmc}.
MotionDiffuse~\cite{zhang2024motiondiffuse}, MDM~\cite{mdm}, and MLD~\cite{chen2023mld} have successfully applied diffusion models to this task, yielding promising results. 
Meanwhile, T2M-GPT~\cite{t2mgpt} and MoMask~\cite{guo2024momask} quantized motions and employed transformer networks for high-quality and faithful motion generation. 
Despite their success, these methods typically rely on coarse textual annotations that often overlook fine-grained action details. 
To address this problem, several works~\cite{kalakonda2023actiongpt, he2023semanticboost, yazdian2023motionscript, zhang2024finemogen} have explored more detailed motion descriptions.
But these descriptions either misalign with ground-truth motion~\cite{kalakonda2023actiongpt} or neglect crucial temporal information~\cite{he2023semanticboost, yazdian2023motionscript}.
FineMoGen~\cite{zhang2024finemogen} addresses these issues by introducing a tailored network designed for fine-grained spatio-temporal motion generation on their dataset.
In contrast, our model is versatile, supporting far more than the tasks of motion generation conditioned on coarse or detailed texts. 
Furthermore, it is trained on the widely studied public dataset, HumanML3D \cite{humanml3d}, which enables comprehensive and fair comparisons with existing works.
This ensures a rigorous evaluation of its effectiveness and provides a solid benchmark for future research.

\subsection{Human Motion Captioning}
Research on captioning 3D human motions remains relatively limited.
TM2T~\cite{guo2022tm2t} introduced a novel motion representation that compresses motions into concise sequences of discrete variables, which are then translated into natural language using a neural translation network.
MotionGPT~\cite{openmotionlab_motiongpt} and M$^3$GPT~\cite{luo2024m3gpt} treated human motion as a foreign language and leveraged LLMs to translate it into natural language descriptions.
Rather than fine-tuning the entire LLM, MotionLLM~\cite{wu2024motionllm} adopted a task-specific LoRA adapter to generate captions more efficiently.
While these methods achieve competitive results in human motion captioning, they primarily provide coarse semantic descriptions.
In contrast, our approach not only offers coarse semantic descriptions but also delivers fine-grained, detailed captions for given motions.

\subsection{Motion Large Language Models}
Large language models (LLMs)~\cite{touvron2023llama, touvron2023llama2, dubey2024llama3, t5, gpt4,yudong2024dynamic} 
have undergone significant development, demonstrating remarkable performance across a wide range of linguistic tasks. 
Their success stems from massive training datasets and the substantial number of model parameters.
Recently, some works have endeavored to unify the motion into LLMs, such as LLaMA~\cite{touvron2023llama} and T5~\cite{t5}.
MotionGPT~\cite{zhang2024motiongpt} constructed a motion generator controlled by multi-modal conditions, \ie, texts and human poses in key frames.
Beyond motion generation, MotionGPT~\cite{openmotionlab_motiongpt} presented a unified framework for various motion-related tasks, such as text-to-motion generation, motion captioning, and motion prediction.
MotionLLM~\cite{wu2024motionllm} extended these efforts from single-human to multi-human motion generation, 
while M$^3$GPT~\cite{luo2024m3gpt} introduced music as an additional modality, further enriching the association between text and motion. 
However, existing approaches primarily focus on modeling the relationship between motion and coarse-grained textual descriptions. 
In contrast, our work pioneers fine-grained motion generation and understanding, and explores motion-related tasks across multiple levels of granularity.



\section{Proposed Method}
\label{sec:method}

\begin{figure*}[!t]
\begin{center}
\includegraphics[width=1.0\linewidth]{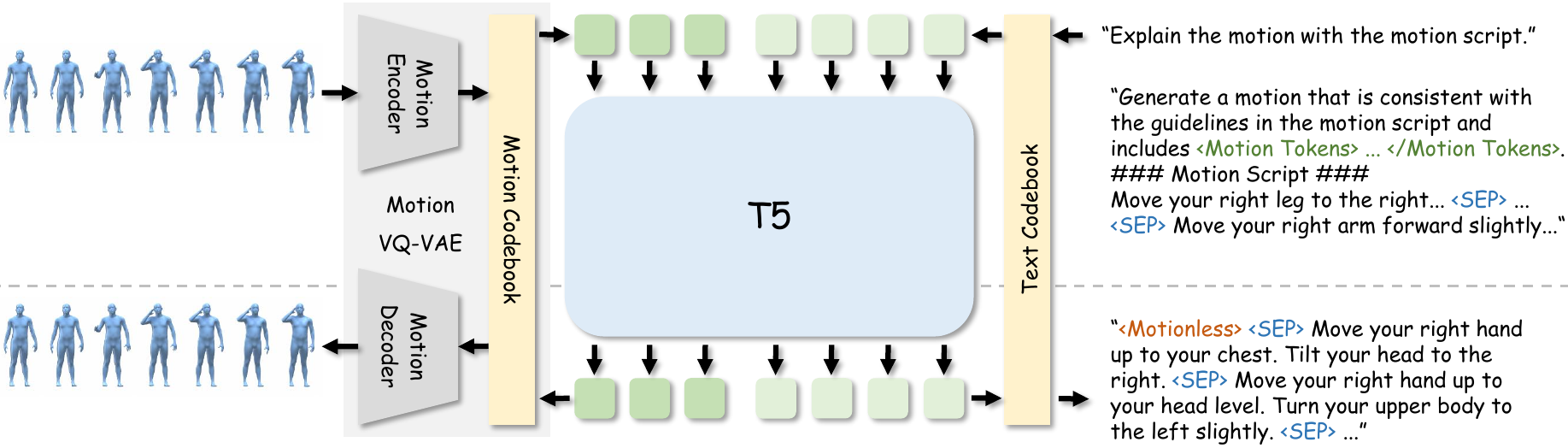}
\end{center}
\setlength{\abovecaptionskip}{-0.3em}  
\caption{
    \textbf{Overview of our MG-MotionLLM.}
    It consists of a motion VQ-VAE and a T5-based motion-aware language model. 
}
\label{fig:fine_motion_llm}
\end{figure*}

To enhance the comprehension and generation of motion-relevant modalities across various granularities, we propose a unified motion-language framework named MG-MotionLLM. 
As shown in Fig.~\ref{fig:fine_motion_llm}, MG-MotionLLM consists of a Motion VQ-VAE (Sec.~\ref{sec:motion_vqvae}) responsible for conversion between the raw motion data and discrete motion tokens, and a motion-aware language model (Sec.~\ref{sec:motion_llm}) to align motion tokens with corresponding textual descriptions. 
Furthermore, a two-stage training scheme, which includes a Granularity-Synergy Pre-training stage and a Task-Specific Instruction Tuning stage, is adopted to facilitate effective motion-aware language model learning across different granularities (Sec.~\ref{sec:training_strategies}).

\subsection{Motion VQ-VAE}
\label{sec:motion_vqvae}
We utilize the motion VQ-VAE in T2M-GPT~\cite{t2mgpt} to represent motions as discrete tokens.
It contains an encoder $\mathcal{E}$, a decoder $\mathcal{D}$, and a learnable codebook $\bm{B}=\{\bm{b}_k\}^{K}_{k=1}$, where $K$ is the size of the codebook. 
Given a $T$-frame motion sequence $\bm{M} = [\bm{m}_1, \bm{m}_2, \dots, \bm{m}_T]$ with $\bm{m}_t \in \mathbb{R}^d$, the encoder $\mathcal{E}$ maps it into a sequence of latent features $\bm{Z} = \mathcal{E}(\bm{M})$ with $\bm{Z} = [\bm{z}_1, \bm{z}_2, \dots, \bm{z}_{T/l}]$ and $\bm{z}_i \in \mathbb{R}^{d_c}$, where $l$ represents the temporal downsampling rate of the encoder $\mathcal{E}$. 
Then, these latent features are transformed into a sequence of motion tokens $\bm{C} = [\bm{c}_1, \bm{c}_2, \dots, \bm{c}_{T/l}]$, 
in which
\begin{equation}
    \bm{c}_i=\underset{\bm{b}_k \in \bm{B}}{\arg\min }\left\|\bm{z}_i-\bm{b}_k\right\|_2 .
\end{equation}
With a sequence of motion tokens $\bm{C}$, 
we first project $C$ back to the corresponding codebook elements $\hat{\bm{Z}} = [\hat{\bm{z}}_1, \hat{\bm{z}}_2, \dots, \hat{\bm{z}}_{T/l}]$ with $\hat{\bm{z}}_i = \bm{b}_{c_i}$. Then, the decoder $\mathcal{D}$ reconstructs $\hat{\bm{Z}}$ into a motion sequence 
$\hat{\bm{M}} = \mathcal{D}(\hat{\bm{Z}}) = [\hat{\bm{m}}_1, \hat{\bm{m}}_2, \dots, \hat{\bm{m}}_T]$. 
The motion VQ-VAE is trained and optimized by three loss functions: the reconstruction loss, the embedding loss, and the commitment loss, 
\ie,
\begin{equation}
    \mathcal{L}_{\text{VQVAE}}=
    \|\bm{M}-\hat{\bm{M}}\|_2
    +\|\mathcal{F}_{\text{SG}}(\bm{Z})-\hat{\bm{Z}}\|_2
    +\beta\|\bm{Z}-\mathcal{F}_{\text{SG}}(\hat{\bm{Z}})\|_2, 
    \end{equation}
where $\beta$ is a hyper-parameter for the commitment loss and $\mathcal{F}_{\text{SG}}(\cdot)$ is the stop-gradient operator.

As shown in Fig.~\ref{fig:fine_motion_llm}, with a learned motion VQ-VAE, motion sequence can be easily mapped into discrete motion tokens by the encoder $\mathcal{E}$ and the codebook $\bm{B}$. 
Conversely, motion tokens in the output of the large language models can be recovered into motion sequences by the decoder $\mathcal{D}$ and the codebook $\bm{B}$, to
facilitate the integration of motion and language for various motion-related tasks.
When the tokenizers are trained, they are frozen for further usage.

\rowcolors{2}{white}{gray!20}
\renewcommand{\arraystretch}{1.1}

\begin{table*}[!b]
\centering
\footnotesize
\setlength{\tabcolsep}{5pt}
\begin{tabular}{m{4.5cm}m{8cm}m{4cm}}

\toprule[1pt]

\textbf{Task} & \textbf{Input} & \textbf{Output} \\

\midrule[1pt]

Text-to-Motion
& Give me a motion that represents the idea of [caption]. 
& [motion] \\

Motion-to-Text
& Describe the action depicted in [motion] using words. 
& [caption] \\

Motion-to-Motion Script
& Describe the movement pattern in [motion] with the motion script. 
& \#\#\# Motion Script \#\#\# [motion script] \\

(Motion Script, Snippet Motion Script)-to-Time
& Find the start and end duration of the snippet of the motion script in the whole motion script. \#\#\# Whole Motion Script \#\#\# [motion script] \#\#\# Snippet Motion Script \#\#\# [snippet motion script]
& [time] \\

\midrule

(Text, Motion Script, Tail Motion)-to-Motion
& Craft a motion that embodies the motion summary, adheres to the motion script, and ends with [tail motion]. \#\#\# Motion Summary \#\#\# [caption] \#\#\# Motion Script \#\#\# [motion script]
& [motion] \\

(Time, Motion Script)-to-Snippet Motion Script
& Detail [time] in the scope of the whole motion script. \#\#\# Whole Motion Script \#\#\# [motion script]
& \#\#\# [time]'s Motion Script \#\#\# [snippet motion script] \\

(Motion, Snippet Motion)-to-Time
& Detail the timing for [snippet motion] in the context of [motion]. 
& [time] \\

\midrule

(Text, Motion Script, Time, Random Motions)-to-Motion
& Provide a [time] motion segment with [random motions], based on a gesture that matches the motion summary and complies with the motion script. \#\#\# Motion Summary \#\#\# [caption] \#\#\# Motion Script \#\#\# [motion script]
& [motion] \\

(Motion, Time)-to-Motion Script
& Describe the action in [motion] for [time] with the motion script. 
& \#\#\# Motion Script \#\#\# [motion script] \\

(Motion, Motion Script)-to-Time
& Specify the timeframe for the snippet of the motion script within [motion]. \#\#\# Motion Script \#\#\# [snippet motion script]
& [time] \\

\bottomrule[1pt]

\end{tabular}
\caption{
    \textbf{Examples of prompt templates} for various tasks used during training, many of which are newly developed. 
    Additional prompt examples and a comprehensive list of tasks are available in the Appendix. 
    We categorize the tasks based on the number of information types included in the input.
}
\label{table: tasks_simplified}
\end{table*}

\rowcolors{2}{}{}
\renewcommand{\arraystretch}{1.0}

\subsection{Motion-Aware Language Model}
\label{sec:motion_llm}

To model text and human motion together, we extend the original text vocabulary $\bm{V}_t=\{\bm{v}_t^i\}_{i=1}^{K_t}$ to incorporate motion vocabulary, $\bm{V}_m=\{\bm{v}_m^i\}_{i=1}^{K_m}$, 
which is composed of the indices of these motion tokens from the codebook $\bm{B}$.
Besides, we include some special tokens $\bm{V}_s$, such as \textcolor{magenta}{\textless Motion Tokens\textgreater} and \textcolor{magenta}{\textless /Motion Tokens\textgreater} for indicating the start and end of the motion token sequences,
\textcolor{cyan}{\textless SEP\textgreater} for separating detailed descriptions of each snippet (\ie, a short segment of the motion sequence),
\textcolor{orange}{\textless Motionless\textgreater} for indicating no significant BPM happened in a snippet.
To sum up, the new unified vocabulary of our LLM model is $\bm{V} = \{ \bm{V}_t, \bm{V}_m, \bm{V}_s \}$, helping formulate diverse motion-related tasks in a general format.
As a result, our MG-MotionLLM can comprehend and generate diverse motion-related sequences using a single model.

To tackle the conditioned generation task, we follow MotionGPT~\cite{openmotionlab_motiongpt} and adopt T5~\cite{t5} as the LLM to effectively map the input into the corresponding output.
Specifically, the input is a sequence of tokens $\bm{X}_\text{in} = \{\bm{v}_\text{in}^i \}_{i=1}^{N_\text{in}}$ as an input, where $\bm{v}_\text{in} \in \bm{V}$ and $N_\text{in}$ is the number of input tokens.  
And the output is represented as $\bm{X}_\text{out} = \{ \bm{v}_\text{out}^i \}_{i=1}^{N_\text{out}}$, where $\bm{v}_\text{out} \in \bm{V}$ and $N_\text{out}$ denotes the length of the output.
As depicted in Fig.~\ref{fig:fine_motion_llm}, the learning process of our LLM can be considered as the next motion token prediction task, based on the given input tokens and the preceding output tokens.
The model is then trained by minimizing the cross-entropy loss between the predicted output tokens and the ground-truth tokens, \ie,
\begin{equation}
    \mathcal{L}_{\text{CE}} = -\sum_{i=1}^{N_\text{out}} log(P(\bm{v}_\text{out}^i \mid \bm{X}_\text{in}, \bm{v}_\text{out}^{j}, \theta_\text{LLM})), j<i.
\end{equation}

After sufficient training, MG-MotionLLM is able to capture the relationship between language and motions, and generate precise answers for given instructions.
During inference, our model samples potential output tokens based on the input token sequence, appends this newly predicted token after the input token sequence, and uses this extended token sequence as the input for subsequent predictions.
The generation process iterates until the model encounters an end token or the output token limit is reached.

\subsection{Training Scheme}
\label{sec:training_strategies}
The training process for the motion-aware language model involves two stages. 
The first stage, Granularity-Synergy Pre-training, aims to align motion with text across multiple granularities,
emphatically establish connections between motions and fine-grained textual descriptions,
and facilitate collaboration among motion-relevant tasks in different granularities. 
The second stage is Task-Specific Instruction Tuning, aiming at further enhancing the model’s capability in specific tasks.

\vspace{0.5em}
\noindent\textbf{Granularity-Synergy Pre-training.}
\label{sec:all_tasks}
Aligning motions with coarse texts is relatively straightforward. 
However, building such connections with detailed textual descriptions poses a significant challenge. 
This phenomenon is observed in our (Text + Detailed Text)-to-Motion task, as shown in Tab.~\ref{table: ablation_training_strategy}, where the result is lower than that of Text-to-Motion, which synthesizes motions conditioned on the coarse captions alone. 
We attribute the inferior performance
to the excessive length of detailed textual descriptions for entire motion sequences, some exceeding 1,000 tokens. 
The vast amount of information makes it difficult to directly extract patterns and match them with concise motion tokens (which contain no more than 50 tokens).
To address this, we dive into short motion segments (composed of several adjacent snippets) and their corresponding detailed descriptions to establish relationships between motions and detailed text, beginning with simpler cases. 
For instance, we ask the model to localize the temporal boundaries of a motion snippet within the complete motion sequence using its detailed text descriptions.
Additionally, we propose to train the model with motion-relevant tasks at various granularities to facilitate mutual reinforcement. 
The coarse-grained tasks support fine-grained tasks in capturing semantic meaning, 
while the fine-grained tasks enhance the model's understanding of motion details for the coarse-grained tasks.

For multi-granular training, we pretrain the model with a total of \textbf{28} distinct motion-relevant tasks, 
including 12 existing classical coarse-grained tasks and 16 newly proposed fine-grained ones.
We summarize the information involved into three types:
(1) textual descriptions (including both coarse caption and detailed texts),
(2) temporal information,
and (3) motion data.
Due to the page limit, we present a subset of these tasks in Tab~\ref{table: tasks_simplified}, which is organized according to the number of information types used in the input.
The upper, middle, and bottom blocks correspond to tasks using one, two, and three types of input information, respectively.
The complete list of all tasks involved in the training process of our MG-MotionLLM can be found in the Appendix.
Notably, motions with different meanings may share the same detailed text descriptions. 
For example, both `walking' and `running' involve alternating the front and back movement of the legs.
Therefore, we exclude tasks that require motion generation when the textual description contains only detailed descriptions.

\vspace{0.5em}
\noindent\textbf{Task-Specific Instruction Tuning.}
\label{sec:task_specific}
After pretraining, our MG-MotionLLM effectively captures the relationships between motions and texts at different granularities.
However, since the model is pretrained on a large number of tasks, each receiving only a relatively small portion of the total training samples (approximately 1/30), it requires fine-tuning on a specific task to achieve optimal adaptation.
To further enhance the performance of our model on specific tasks, we propose to continue the training of the model with task-specific instruction data.


\section{Experimental Results}
\label{sec:experiments}

In this section, we evaluate the performance of our MG-MotionLLM across multiple motion-relevant tasks. 
The datasets, evaluation metrics, and implementation details are introduced in Sec.~\ref{sec:exp_setup}. 
In Sec.~\ref{sec:sota}, we compare our model with other state-of-the-art works on the classical tasks, \ie, text-to-motion and motion-to-text. 
We also present quantitative results for the newly proposed fine-grained motion-relevant tasks: motion-to-detailed text.
In addition, qualitative results demonstrate several novel applications of our model.
Finally, we assess the effectiveness of our contributions through ablation studies in Sec.~\ref{sec:ablation}.

\subsection{Experimental Setup}
\label{sec:exp_setup}

\noindent\textbf{Datasets.}
Our evaluations are conducted on two motion-language datasets.
The first is HumanML3D~\cite{humanml3d}, one of the largest motion-text pair datasets. 
It contains 14,616 motion sequences from AMASS~\cite{mahmood2019amass} and HumanAct12~\cite{humanact12}, along with 44,970 sequence-level textual descriptions.
These descriptions capture the semantic meaning of each motion, making HumanML3D widely used for coarse-grained motion-related tasks.
The second is FineMotion~\cite{Wu_FineMotion_Dataset}, an open-source dataset that re-label all motions in HumanML3D with detailed body movement descriptions over time.
Specifically, each motion is segmented into snippets at fixed temporal intervals, resulting in 420,968 snippets.
Each snippet is paired with fine-grained descriptions of body part movements.
Examples and details can be found as \#\#\# Motion Script \#\#\# part in Fig.~\ref{fig:intro} and the Appendix.
This dataset is used for fine-grained motion-related tasks.

\begin{table*}[!t]
\begin{center}
\footnotesize
\setlength{\tabcolsep}{5pt} 
\begin{tabular}{clccccccc}
    \toprule[1pt]
    
    \multirow{2}{*}{Types} & \multirow{2}{*}{Methods} & \multicolumn{3}{c}{R-Precision $\uparrow$} & \multirow{2}{*}{FID $\downarrow$} & \multirow{2}{*}{MM-Dist $\downarrow$}  & \multirow{2}{*}{Diversity $\uparrow$}  & \multirow{2}{*}{MModality $\uparrow$} \\ 

    \cmidrule(r){3-5}

    & & Top-1 & Top-2 & Top-3 & & & &  \\

    \midrule[0.5pt]

    & Real motion & 0.511$^{\pm.003}$ & 0.703$^{\pm.003}$ & 0.797$^{\pm.002}$ & 0.002$^{\pm.000}$ & 2.974$^{\pm.008}$ & 9.503$^{\pm.065}$ & - \\
    
    \midrule[0.5pt]
    
    \multirow{8}{*}{\makecell{Motion Gen. \\ Only}} & TEMOS~\cite{petrovich2022temos} $_{\text{ECCV'22}}$ & 0.424$^{\pm.002}$ & 0.612$^{\pm.002}$ & 0.722$^{\pm.002}$ & 3.734$^{\pm.028}$ & 3.703$^{\pm.008}$ & 8.973$^{\pm.071}$ & 0.368$^{\pm.018}$ \\
    & TM2T~\cite{guo2022tm2t} $_{\text{ECCV'22}}$ & 0.424$^{\pm.003}$ & 0.618$^{\pm.003}$ & 0.729$^{\pm.002}$ & 1.501$^{\pm.017}$ & 3.467$^{\pm.011}$ & 8.589$^{\pm.076}$ & 2.424$^{\pm.093}$ \\
    & Guo et al.\cite{humanml3d} $_{\text{CVPR'22}}$ & 0.455$^{\pm.003}$ & 0.636$^{\pm.003}$ & 0.736$^{\pm.002}$ & 1.087$^{\pm.021}$ & 3.347$^{\pm.008}$ & 9.175$^{\pm.083}$ & 2.219$^{\pm.074}$ \\
    & MotionDiffuse~\cite{zhang2024motiondiffuse} $_{\text{TPAMI'24}}$ & 0.491$^{\pm.001}$ & 0.681$^{\pm.001}$ & 0.782$^{\pm.001}$ & 0.630$^{\pm.001}$ & 3.113$^{\pm.001}$ & 9.410$^{\pm.049}$ & 1.553$^{\pm.042}$ \\
    & T2M-GPT~\cite{t2mgpt} $_{\text{CVPR'23}}$ & 0.492$^{\pm.003}$ & 0.679$^{\pm.002}$ & 0.775$^{\pm.002}$ & 0.141$^{\pm.004}$ & 3.121$^{\pm.009}$ & \textbf{9.722}$^{\pm.082}$ & 1.831$^{\pm.048}$ \\
    & FineMoGen~\cite{zhang2024finemogen} $_{\text{NeurIPS'24}}$ & 0.504$^{\pm.002}$ & 0.690$^{\pm.002}$ & 0.784$^{\pm.002}$ & 0.151$^{\pm.008}$ & 2.998$^{\pm.008}$ & 9.263$^{\pm.094}$ & \textbf{2.696}$^{\pm.079}$ \\
    & MoMask~\cite{guo2024momask} $_{\text{CVPR'24}}$ & \textbf{0.521}$^{\pm.002}$ & \textbf{0.713}$^{\pm.002}$ & \textbf{0.807}$^{\pm.002}$ & \textbf{0.045}$^{\pm.002}$ & \textbf{2.958}$^{\pm.008}$ & - & 1.241$^{\pm.040}$ \\

     & MotionGPT~\cite{zhang2024motiongpt} $_{\text{AAAI'24}}$ & - & - & - & 0.567\hspace{2.4em} & 3.775\hspace{2.4em} & 9.006\hspace{2.4em} & - \\

    \midrule[0.5pt]

    \multirow{2}{*}{\makecell{Unified Motion \\ Gen. and Und.}} & MotionGPT~\cite{openmotionlab_motiongpt} $_{\text{NeurIPS'23}}$ & 0.492$^{\pm.003}$ & 0.681$^{\pm.003}$ & 0.778$^{\pm.002}$ & \textbf{0.232}$^{\pm.008}$ & 3.096$^{\pm.008}$ & 9.528$^{\pm.071}$ & 2.008$^{\pm.084}$ \\

    & \textbf{MG-MotionLLM (Ours)} & \textbf{0.516}$^{\pm.002}$ & \textbf{0.706}$^{\pm.002}$ & \textbf{0.802}$^{\pm.003}$ & 0.303$^{\pm.010}$ & \textbf{2.952}$^{\pm.009}$ & \textbf{9.960}$^{\pm.073}$ & \textbf{2.125}$^{\pm.159}$ \\

    \bottomrule[1pt]
    
\end{tabular}
\setlength{\abovecaptionskip}{5pt} 
\caption{
    Comparison of \textbf{motion generation} on the HumanML3D~\cite{humanml3d} test set.
    We group the existing methods into ones that solely focus on motion generation (\textit{Motion Gen. Only}) and ones that unify motion understanding and generation tasks (\textit{Unified Motion Gen. and Und.}).
    \textbf{Bold} results refer to the best ones in each block.
    Results showed that our MG-MotionLLM achieves state-of-the-art performance.
} \label{table: humanml3d_t2m}
\vspace{-15pt}
\end{center}
\end{table*}

\begin{table}[!t]
\begin{center}
\footnotesize
\setlength{\tabcolsep}{2.8pt} 
\begin{tabular}{lcccccc}
    \toprule[1pt]
    
    \multirow{2}{*}{Methods} & \multicolumn{2}{c}{R-Precision $\uparrow$} & \multirow{2}{*}{MM-Dist $\downarrow$}  & \multirow{2}{*}{Bleu@4 $\uparrow$} & \multirow{2}{*}{BertScore $\uparrow$} \\ 

    \cmidrule(r){2-3}

     & Top-1 & Top-3 &  \\

    \midrule[0.5pt]

    Real & 0.523 & 0.828 & 2.901 & - & - \\

    \midrule[0.5pt]
    
    TM2T~\cite{guo2022tm2t} & 0.516 & 0.823 & 2.935 & 7.00 & 32.2 \\
    MotionGPT~\cite{openmotionlab_motiongpt} & \underline{0.543} & \underline{0.827} & \underline{2.821} & \textbf{12.47} & \underline{32.4} \\

    
    \textbf{MG-MotionLLM} & \textbf{0.592} & \textbf{0.866} & \textbf{2.581} & \underline{8.06} & \textbf{36.7}  \\

    \bottomrule[1pt]
    
\end{tabular}
\setlength{\abovecaptionskip}{5pt} 
\caption{
    \textbf{Motion captioning} results on the HumanML3D~\cite{humanml3d} test set.
    We \textbf{bold} the best result and \underline{underline} the second-best one.
    Our model exceeds the previous methods on most metrics.
} \label{table: humanml3d_m2t}
\vspace{-2em}
\end{center}
\end{table}

\vspace{0.5em}
\noindent\textbf{Evaluation Metrics.}
The evaluation metrics are summarized into two categories:
(1) Motion Quality Assessment: 
We use metrics aligned with prior research~\cite{t2mgpt, humanml3d} to evaluate the realism and diversity of generated motion. 
These metrics include Frechet Inception Distance (FID), Multi-modal Distance (MM-Dist), R-Precision (Top-1/2/3 motion-to-text retrieval accuracy), and Diversity.
(2) Text Quality Assessment: 
Following~\cite{guo2022tm2t}, we evaluate text using linguistic metrics like BLEU~\cite{papineni2002bleu}, ROUGE~\cite{rouge2004package}, and BERTScore~\cite{zhang2019bertscore}.
Apart from assessing the generated detailed text from the whole sequence level, we also evaluate it from the snippet level.
Besides, R-Precision from~\cite{humanml3d} is employed to assess motion-text matching accuracy. 
Finally, MM-Dist measures the text-motion alignment.

\vspace{0.5em}
\noindent\textbf{Implementation Details.}
For fair comparisons and to facilitate further research in the field, we follow~\cite{humanml3d} and adopt a consistent motion representation with a dimensionality of 263.
All snippet descriptions within a motion sequence are connected into a single long text via the special token \textcolor{cyan}{\textless SEP\textgreater}, 
with empty snippet descriptions replaced by \textcolor{orange}{\textless Motionless\textgreater}.
For the motion VQ-VAE, we follow the network structure and training strategy in~\cite{t2mgpt}.
As for the motion-aware language model, we use three sizes of pretrained T5~\cite{t5} models, \ie, small, base, and large.
All models are trained using the AdamW optimizer.
Pretraining is done with a learning rate of $2 \times 10^\text{-4}$, followed by fine-tuning for specific tasks at $10^\text{-4}$.
The base T5 model uses a batch size of 16 and undergoes 300K iterations during pretraining and 300K during task-specific instruction tuning. 
All experiments are conducted on a single Tesla A100 80G GPU. 
Additional details are provided in the Appendix.


\subsection{Results of Motion-relevant Tasks}
\label{sec:sota}

\noindent\textbf{Comparisons on Text-to-Motion.} 
The text-to-motion task involves generating human motion sequences based on coarse captions. 
We compare our MG-MotionLLM with other state-of-the-art methods on the HumanML3D dataset using the same metrics as in~\cite{humanml3d}. 
The results are reported with a 95\% confidence interval, obtained from 20 repeated trials. 
Tab.~\ref{table: humanml3d_t2m} summarizes the comparison results, 
where we divide them into two groups: 
those that solely focus on motion generation and 
those that integrate motion understanding with generation tasks.
It can be observed that our MG-MotionLLM surpasses MotionGPT \cite{openmotionlab_motiongpt}, which also aims to unify multiple motion-relevant tasks, 
while achieving competitive performance with methods tailored to motion generation across most metrics.
Qualitative results can be found in the Appendix.

\noindent\textbf{Comparisons on Motion-to-Text.} 
The motion-to-text task involves generating coarse textual descriptions from motion sequences. 
We compare our MG-MotionLLM with TM2T~\cite{guo2022tm2t} and MotionGPT~\cite{openmotionlab_motiongpt} in Tab.~\ref{table: humanml3d_m2t}.
The compared results are taken directly from~\cite{openmotionlab_motiongpt}.
Following~\cite{openmotionlab_motiongpt}, we use original ground-truth text descriptions for a more precise assessment, rather than preprocessing them to ignore grammatical tense and plural forms as in~\cite{guo2022tm2t}.
The results show that our MG-MotionLLM significantly outperforms previous models in captioning motion sequences, particularly in metrics that assess semantic content like BertScore and text-motion alignment like R-Precision and MM-Dist.

\begin{figure*}[!t]
\begin{center}
\includegraphics[width=1.0\linewidth]{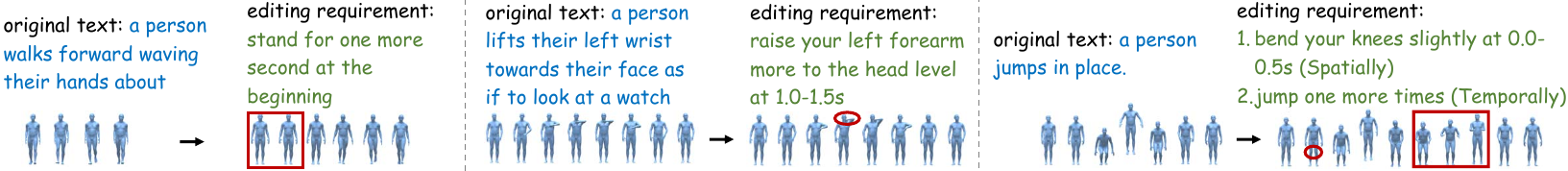}
\end{center}
\setlength{\abovecaptionskip}{-0.3em}  
\caption{
    \textbf{Text-Driven Fine-grained Motion Editing Examples.} 
    We display some examples of temporal editing (\textit{left}), spatial editing (\textit{middle}), and spatial-temporal editing (\textit{right}).
}
\label{fig:editing_examples}
\end{figure*}

\begin{figure*}[!t]
\begin{center}
\includegraphics[width=1.0\linewidth]{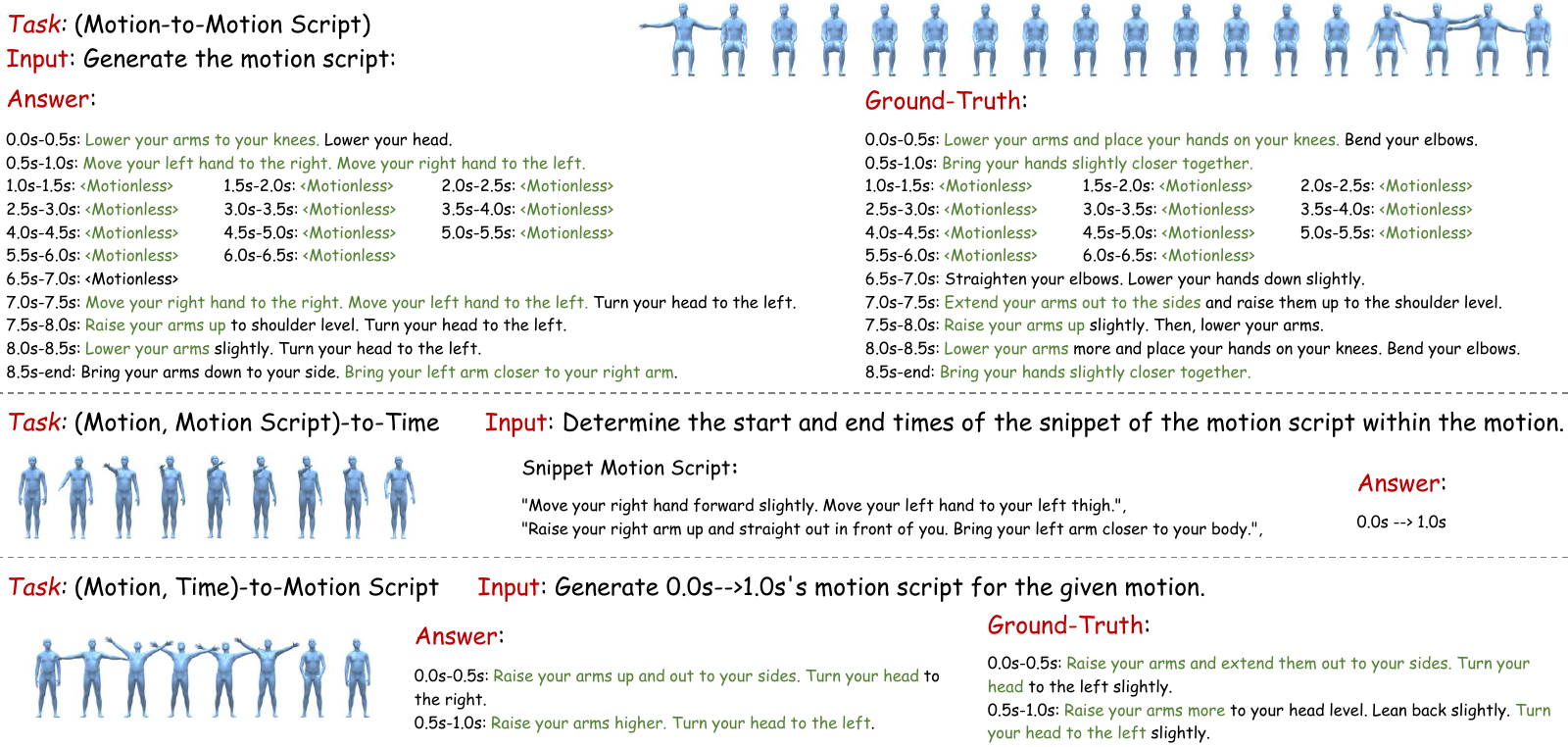}
\end{center}
\setlength{\abovecaptionskip}{-0.3em}  
\caption{
    \textbf{Other Novel Applications.} 
    We display some examples of brand-new tasks: fine-grained captioning of both whole (\textit{up}) and partial (\textit{bottom}) motion sequences, and motion localization via fine-grained textual description (\textit{middle}).
}
\label{fig:other_app_examples}
\end{figure*}

\begin{table}[!t]
\begin{center}
\footnotesize
\setlength{\tabcolsep}{1.8pt} 
\begin{tabular}{lcccccc}
    \toprule[1pt]
    
     Level & Size & Bleu@1 $\uparrow$ & Bleu@4 $\uparrow$  & Bleu@7 $\uparrow$ & Rouge $\uparrow$ & BertScore $\uparrow$ \\

    \midrule[0.5pt]
    
    \multirow{3}{*}{Sequence} & Small & 79.02 & 63.61 & 51.02 & 64.6 & 50.2 \\

    & Base & \textbf{81.78} & \textbf{66.44} & \textbf{53.75} & \underline{65.5} & \underline{52.3} \\
    
    & Large & \underline{81.73} & \underline{66.29} & \underline{53.56} & \textbf{65.8} & \textbf{52.4} \\

    \midrule[0.5pt]
    
    \multirow{3}{*}{Snippet} & Small & 65.33 & 44.93 & 31.99 & 57.8 & 47.4 \\

    & Base & \underline{66.57} & \underline{46.94} & \underline{34.43} & \underline{59.6} & \underline{49.8} \\
    
    & Large & \textbf{68.25} & \textbf{49.26} & \textbf{37.07} & \textbf{61.4} & \textbf{52.2} \\
    
    \bottomrule[1pt]
    
\end{tabular}
\setlength{\abovecaptionskip}{5pt} 
\caption{
    Results of \textbf{Motion-to-Detailed Text} at two levels on the FineMotion test set.
    Results show that the larger model size yields better performance on most metrics.
} \label{table: m2dt}
\vspace{-2em}
\end{center}
\end{table}

\noindent\textbf{Results on Motion-to-Detailed Text.} 
The motion-to-detailed text task involves generating detailed textual descriptions from motion sequences.
We evaluate this task at two levels: sequence level and snippet level.
For the sequence level, we use linguistic metrics from natural language processing to compare the predicted detailed description of the entire motion sequence against the ground truth.
As for the snippet level, we first split the predicted and ground-truth detailed text for the whole motion sequence into descriptions for snippet via the special token \textcolor{cyan}{\textless SEP\textgreater}, and then we assess them snippet by snippet.
As shown in Tab.~\ref{table: m2dt}, this task is more challenging than motion-to-text or text-to-motion tasks, and the performance generally improves with larger models. 
More discussions are given in the Appendix.

\noindent\textbf{Novel Applications.}
We display some novel application examples of our MG-MotionLLM, which leverages the Granularity-Synergy Pre-trained T5-Base model for visualization.
Previous motion editing methods~\cite{kim2023flame, mdm, tevet2022motionclip, tseng2023edge} rely on coarse texts for editing, which limits their ability to precisely control the timing of specific body movements and their duration. 
Instead, our model enables precise human motion editing through detailed text. 
Specifically, users can employ our model to synthesize a preliminary motion given a coarse caption.
Then, our model can output the detailed body part movement descriptions over time for this preliminary motion.
Next, users can precisely edit the returned detailed text to meet their fine-grained editing requirements.
Finally, our model can generate the edited motions via the (text + detailed text)-to-motion task, where the detailed text refers to the edited version. 
Notably, all these steps can be accomplished by using a single model. 
Fig.~\ref{fig:editing_examples} presents examples of this innovative application, and it demonstrates the model’s ability to precisely edit human motion along spatial, temporal, and spatiotemporal dimensions.
Additionally, our model supports a wide range of novel fine-grained motion-related applications, such as fine-grained captioning of both the whole and the part of motion sequence, as well as motion localization through detailed textual descriptions as shown in Fig.~\ref{fig:other_app_examples}.
The Granularity-Synergy Pre-training stage enhances the model's ability to handle all these tasks within a unified framework by using various prompting templates. 
Refer to the Appendix for more novel applications of our model.


\subsection{Ablation Study}
\label{sec:ablation}

\begin{table*}[!t]
\begin{center}
\footnotesize
\setlength{\tabcolsep}{3pt} 
\begin{tabular}{cccccccccccccc}
    \toprule[1pt]
    
    \multirow{2}{*}{\makecell{Pretraining \\ Granularity}} & \multirow{2}{*}{Ins.} & \multicolumn{3}{c}{Text-to-Motion} & \multicolumn{3}{c}{(Text + Detailed Text)-to-Motion} & \multicolumn{3}{c}{Motion-to-Text}  & \multicolumn{3}{c}{Motion-to-Detailed Text} \\ 

    \cmidrule(r){3-5}\cmidrule(r){6-8}\cmidrule(r){9-11}\cmidrule(r){12-14}

     &  & Top-3 $\uparrow$ & FID $\downarrow$ & DIV $\uparrow$ & Top-3 $\uparrow$ & FID $\downarrow$ & DIV $\uparrow$ & MMDist $\downarrow$ & Top-1 $\uparrow$ & BertScore $\uparrow$ & Bleu@4 $\uparrow$ & Rouge $\uparrow$ & BertScore $\uparrow$  \\
    
    \midrule[0.5pt]

    \multicolumn{2}{c}{\textcolor{gray}{Real}} & 0.797 & 0.002 & 9.503 & 0.797 & 0.002 & 9.503 & 2.901 & 0.523 & - & - & - & - \\

    \midrule[0.5pt]
    
     - & \checkmark & \underline{0.773} & \underline{0.485} & 9.862 & 0.750 & \textbf{0.148} & \textbf{9.790} & 2.969 & \underline{0.516} & 22.7 & 62.34 & 63.6 & \underline{50.5} \\

     Coarse & - & 0.725 & 0.862 & 9.877 & - & - & - & 3.433 & 0.431 & \underline{37.5} & - & - &  \\

     Fine & - & - & - & - & 0.718 & 0.328 & 9.769 & - & - & - & 55.32 & 62.4 & 43.1 \\

     Coarse + Fine & - & 0.767 & 0.540 & \underline{9.953} & \underline{0.771} & 0.313 & 9.739 & \underline{2.942} & 0.514 & \textbf{38.6} & \underline{62.87} & \underline{64.6} & 47.7 \\

     Coarse + Fine & \checkmark & \textbf{0.802} & \textbf{0.303} & \textbf{9.960} & \textbf{0.779} & \underline{0.195} & \underline{9.783} & \textbf{2.581} & \textbf{0.592} & 36.7 & \textbf{66.44} & \textbf{65.5} & \textbf{52.3} \\

    \bottomrule[1pt]
    
\end{tabular}
\setlength{\abovecaptionskip}{5pt} 
\caption{
    Ablation of the proposed \textbf{training scheme} in four motion tasks on HumanML3D~\cite{humanml3d} dataset.
    Results show that the Granularity-Synergy Pre-trained model outperforms models that are directly instruction-tuned on specific tasks, even with relatively few training iterations per task (approximately 1/30). 
    Pretraining the model with tasks spanning multiple granularities promotes mutual enhancement across tasks at each granularity. 
    Further instruction tuning on specific tasks leads to significant performance gains. 
} \label{table: ablation_training_strategy}
\vspace{-1em}
\end{center}
\end{table*}

\begin{table*}[!t]
\begin{center}
\footnotesize
\setlength{\tabcolsep}{4pt} 
\begin{tabular}{lccccccccccccc}
    \toprule[1pt]
    
    \multirow{2}{*}{Size} & \multirow{2}{*}{Param.} & \multicolumn{3}{c}{Text-to-Motion} & \multicolumn{3}{c}{(Text + Detailed Text)-to-Motion} & \multicolumn{3}{c}{Motion-to-Text}  & \multicolumn{3}{c}{Motion-to-Detailed Text} \\ 

    \cmidrule(r){3-5}\cmidrule(r){6-8}\cmidrule(r){9-11}\cmidrule(r){12-14}

    &  & Top-3 $\uparrow$ & FID $\downarrow$ & DIV $\uparrow$ & Top-3 $\uparrow$ & FID $\downarrow$ & DIV $\uparrow$ & MMDist $\downarrow$ & Top-1 $\uparrow$ & BertScore $\uparrow$ & Bleu@4 $\uparrow$ & Rouge $\uparrow$ & BertScore $\uparrow$ \\

    \midrule[0.5pt]

    \multicolumn{2}{c}{\textcolor{gray}{Real}} & 0.797 & 0.002 & 9.503 & 0.797 & 0.002 & 9.503 & 2.901 & 0.523 & - & - & - & - \\
    
    \midrule[0.5pt]
    
    Small & 60M  & 0.781 & 0.455 & 9.881 & \underline{0.767} & \underline{0.239} & 9.740 & 2.641 & 0.571 & \underline{37.2} & 63.61 & 64.6 & 50.2 \\

    Base  & 220M & \textbf{0.802} & \textbf{0.303} & \textbf{9.960} & \textbf{0.779} & \textbf{0.195} & \textbf{9.783} & \textbf{2.581} & \textbf{0.592} & 36.7 & \textbf{66.44} & \underline{65.5} & \underline{52.3} \\

    Large & 770M & \underline{0.782} & \underline{0.401} & \underline{9.954} & 0.763 & 0.281 & \underline{9.762} & \underline{2.615} & \underline{0.573} & \textbf{37.9} & \underline{66.29} & \textbf{65.8} & \textbf{52.4} \\

    \bottomrule[1pt]
    
\end{tabular}
\setlength{\abovecaptionskip}{5pt}
\caption{
    Ablation of different \textbf{model sizes} of MG-MotionLLMs in four motion tasks on HumanML3D~\cite{humanml3d} dataset.
    Similar to MotionGPT~\cite{openmotionlab_motiongpt}, the increasing model size does not yield consistent improvements in all tasks, due to the limited scale of HumanML3D.
} \label{table: ablation_model_size}
\end{center}
\vspace{-1em}
\vspace{-5pt}
\end{table*}

Both the proposed training scheme and the model size influence the performance of our MG-MotionLLM. We hence Ablate them on both coarse- and fine-grained motion-relevant tasks.   
More details are provided in the Appendix.

To evaluate the proposed \textbf{training scheme}, we analyze the impact of each training stage on the base model in Tab.~\ref{table: ablation_training_strategy}. 
The results demonstrate that the Granularity-Synergy Pre-trained model (Row 4) outperforms models that are directly instruction-tuned on specific tasks (Row 1), such as fine-grained motion-relevant tasks and the motion-to-text task, even though the average training iterations per task are relatively low (approximately 1/30). 
Additionally, pretraining the model with tasks spanning multiple granularities promotes mutual enhancement across tasks within each granularity, by contrasting the results in Row 4 against the results in Rows 3 and 2. 
Moreover, further instruction tuning on specific tasks leads to significant performance gains across all four evaluated tasks, as shown in Row 5.

Notably, the results for Top-3 Retrieval Accuracy in the (Text + Detailed Text)-to-Motion task are slightly lower than those for the Text-to-Motion task, as this metric is computed based on the distance between coarse textual embeddings and motion embeddings, both derived from a  network~\cite{humanml3d} pretrained on paired (coarse text, motion) data.
This network is good at capturing global semantic associations between coarse text and corresponding motion features. 
But motion generated from fine-grained text typically contains more detailed and complex characteristics, which may not well correspond to the overall semantics of coarse text, leading to a decline in this metric.
In contrast, motion-only metrics, such as FID and Diversity, indicate that motions generated from both coarse captions and detailed texts more closely resemble the ground truth.
By default, MG-MotionLLM is trained using the proposed training scheme.

As for the \textbf{model size}, we evaluate MG-MotionLLMs at three model sizes: small (60M), base (220M), and large (770M).
Tab.~\ref{table: ablation_model_size} illustrates that the base model achieves better performance compared to the small model.
Besides, consistent with observations from MotionGPT~\cite{openmotionlab_motiongpt}, further increasing model size does not lead to notable performance improvements and can even degrade results, as observed in the (Text + Detailed Text)-to-Motion task for the large model. 
This phenomenon can be attributed to the limited scale of HumanML3D, which comprises only 14,616 motion sequences, a stark contrast to the billions of samples available in language and image datasets.

\section{Conclusion}

In this paper, we introduce MG-MotionLLM, a unified framework for motion understanding and generation across multiple granularities. 
We also propose a comprehensive multi-granularity training scheme that enables the model to establish connections between motions and their corresponding detailed textual descriptions, and foster mutual enhancement among various motion-relevant tasks at different granularities. 
Extensive evaluations across a range of motion-related tasks demonstrate the effectiveness of MG-MotionLLM in both motion comprehension and generation at coarse and fine granularity.
Finally, our MG-MotionLLM investigates a lot of brand-new applications by using one unified framework, offering a fresh perspective in the field.


\clearpage
\newpage

\section{Acknowledgements}
This work was supported by the National Key R\&D Program of China (No. 2024YFF0618403), National Natural Science Foundation of China under Grant 82261138629;  Guangdong Provincial Key Laboratory under Grant 2023B1212060076, and Shenzhen Municipal Science and Technology Innovation Council under Grant JCYJ20220531101412030.

{
    \small
    \bibliographystyle{ieeenat_fullname}
    \bibliography{main}
}

\clearpage
\setcounter{page}{1}

\begin{appendices}
\maketitlesupplementary


\section{More Implementation Details}
\label{sec:more_im_details}

Apart from the MG-MotionLLM with 220M parameters, 
we implement a smaller model with 60M parameters as well as a larger model with 770 million parameters.
The training settings strictly follow MotionGPT \cite{openmotionlab_motiongpt}. 
Refer to Tab.~\ref{tab:appendix_im_details} for more details.
Notably, according to \cite{guo2022tm2t}, motion captioning is typically easier than text-to-motion, so it is necessary to reduce the training iterations when instruction-tuning for the motion captioning task to avoid overfitting.
Specifically, we instruction-tuned the small, base, and large models with 200K, 100K, and 200K on this task, respectively.

\begin{table}[!h]
\footnotesize
\centering
    \begin{tabular}{@{}lccc@{}}
        \toprule[1pt]
        MotionGPT &  Small & Base & Large
        \\ 
        \midrule[0.5pt]
        Backbone & T5-Small & T5-Base & T5-Large \\
        Training Batch Size & 64 &16 &4 
        \\
        Model Size& 60M&220M&770M
        \\
        Pre-training - Iterations & 300K& 300K& 300K
        \\
        Pre-training - Learning Rate & 2e-4& 2e-4& 2e-4
        \\
        Instruction Tuning - Iterations & 200K& 300K& 400K
        \\
        Instruction Tuning - Learning Rate & 1e-4& 1e-4& 1e-4
        \\
        \bottomrule[1pt]
    \end{tabular}%
\caption{Hyperparameters for different MG-MotionLLMs.}
\label{tab:appendix_im_details}
\end{table}


\section{Details of the FineMotion Dataset}

\begin{figure*}[!b]
\begin{center}
\includegraphics[width=1.0\linewidth]{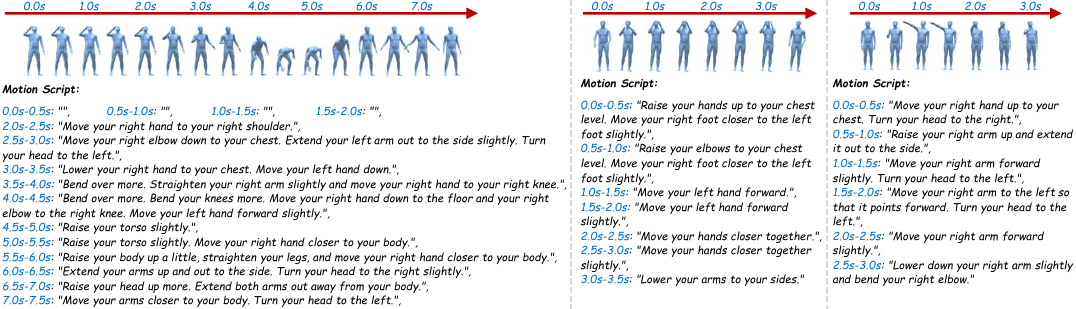}
\end{center}
\vspace{-1em}
\caption{
    Examples of motion scripts for motion sequences in the FineMotion dataset.
}
\label{fig:dataset_examples_SM}
\end{figure*}

The FineMotion~\cite{Wu_FineMotion_Dataset} dataset builds upon the existing text-motion pairs dataset, HumanML3D \cite{humanml3d}, but describes human motions in fine detail both spatially and temporally.
Specifically, each motion sequence is divided into snippets of fixed temporal intervals, resulting in a total of 420,968 motion snippets.
Each snippet is paired with an automatically generated, fine-grained description of body part movements.
In addition, 21,346 motion snippets - covering 5\% of all sequences - are manually annotated.
When manual annotations are available, they are prioritized; otherwise, the automated annotations are used.

The detailed body part movement descriptions for all snippets within a motion sequence can be considered the \textit{motion script} of that sequence.
These extended descriptions are strictly \textbf{aligned} with the motions and explicitly incorporate \textbf{temporal information}, addressing the challenges discussed in Sec.~\ref{sec:related_work_t2m}.
Examples of (motion sequence, motion script) pairs are illustrated in Fig.~\ref{fig:dataset_examples_SM}.


\section{Discussions on Motion-to-Detailed Text}

It is observed that our MG-MotionLLM achieves significantly higher scores on the `Motion-to-Detailed Text' task - despite it being an apparently more challenging task - compared to the `Motion-to-Text' task. 
This can be attributed to the fact that fine-grained descriptions typically follow a consistent structure, such as `verb + body part + direction' (e.g., \textit{move your right arm forward}), whereas coarse descriptions exhibit more varied and less predictable patterns.
The structural consistency of fine-grained descriptions contributes to higher scores on metrics such as BERTScore.


\section{More Detailed Ablation Study}

In this section, we conduct an ablation study to assess the contribution of each task included in the Granularity-Synergy Pre-training stage (see \ref{sec:more_tasks} for definitions and template examples).
Specifically, we pretrain 28 models, each of which excludes one task and is trained with the remaining 27 tasks.
Since different tasks are evaluated using different metrics, it is infeasible to assess the overall performance of these pretrained models across all 28 tasks with a single unified metric.
Therefore, we select three representative tasks for evaluation: \ie, Text-to-Motion, Motion-to-Text, and (Text, Detailed Text)-to-Motion. 
These tasks cover both coarse-grained and fine-grained aspects, as well as both generation and comprehension abilities, and all utilize retrieval accuracy as the evaluation metric. 
Notably, for models `w/o Text-to-Motion', `w/o Motion-to-Text', and `w/o (Text, Detailed Text)-to-Motion', we evaluate them on the remaining two representative tasks.
We report the average Top-1 Retrieval Accuracy across representative tasks.

\begin{figure*}[!t]
\begin{center}
\includegraphics[width=1.0\linewidth]{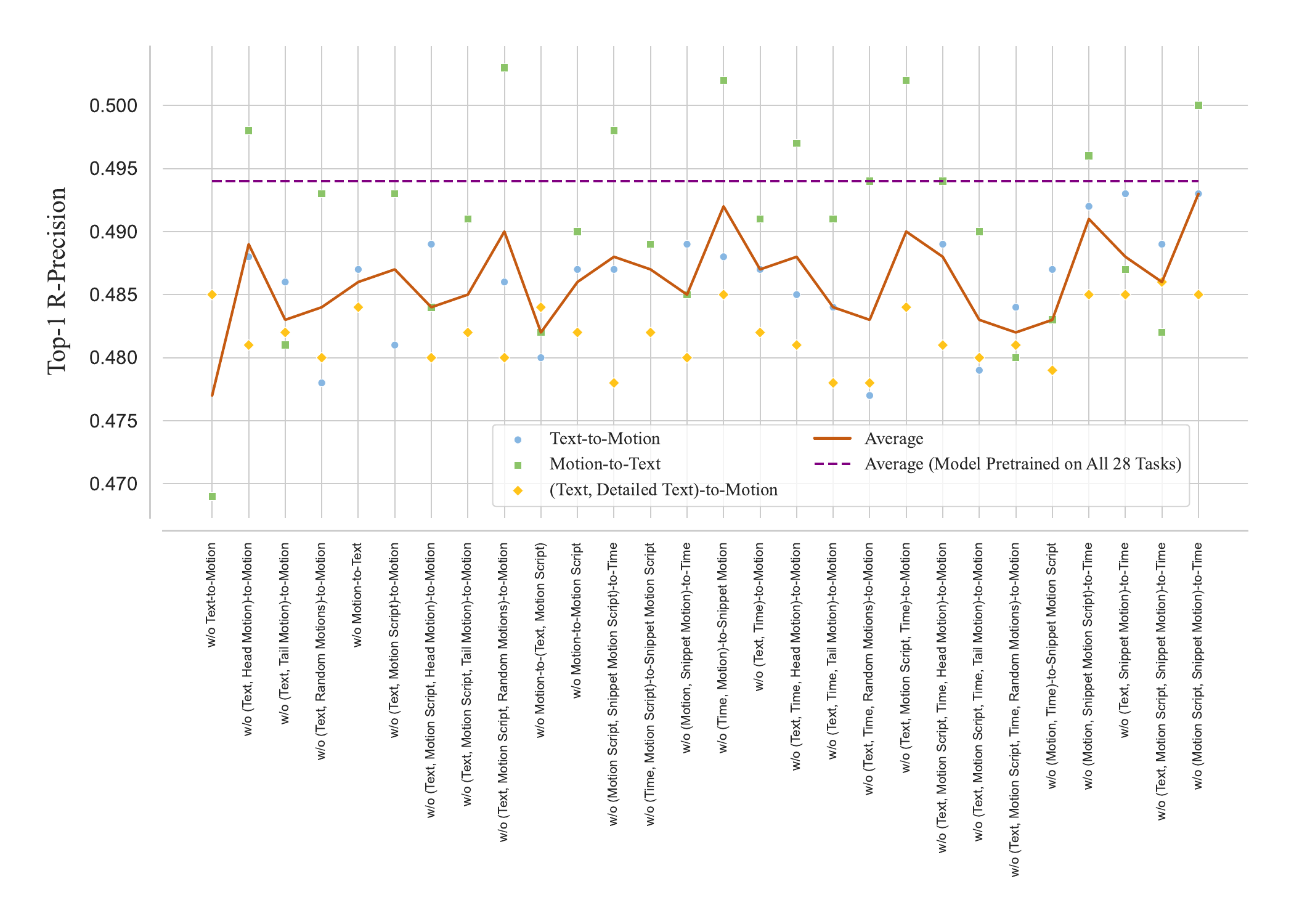}
\end{center}
\caption{
    \textbf{Ablation of all the tasks in the Granularity-Synergy Pre-training stage on the HumanML3D dataset.}
    To assess overall performance, we evaluate three representative tasks, \ie, \textbf{Text-to-Motion}, \textbf{Motion-to-Text}, and \textbf{(Text, Detailed Text)-to-Motion}, which cover both coarse- and fine-grained aspects, including generation and comprehension. 
    Notably, for models `w/o Text-to-Motion', `w/o Motion-to-Text', and `w/o (Text, Detailed Text)-to-Motion', we only evaluate the other two representative tasks.
    All these tasks use retrieval accuracy as the evaluation metric and we report their average Top-1 Retrieval Accuracy.
}
\label{fig:more_ablation}
\end{figure*}

As illustrated in Fig.~\ref{fig:more_ablation}, 
the average performances of all these 28 models pretrained on 27 tasks is consistently lower than that of the model pretrained on all 28 tasks. 
This observation highlights the importance of all tasks in this stage, confirming that their collective contribution is crucial for the comprehensive capability of our MG-MotionLLM in both generating and understanding motion across different granularities.
Among the 28 tasks, text-to-motion, Motion-to-(Text, Motion Script), and (Text, Motion Script, Time, Random Motions)-to-Motion have the greatest, second-greatest, and third-greatest influence on the performance of our MG-MotionLLM, respectively.


\section{Qualitative Results of Text-to-Motion}

\begin{figure*}[!t]
\begin{center}
\includegraphics[width=1.0\linewidth]{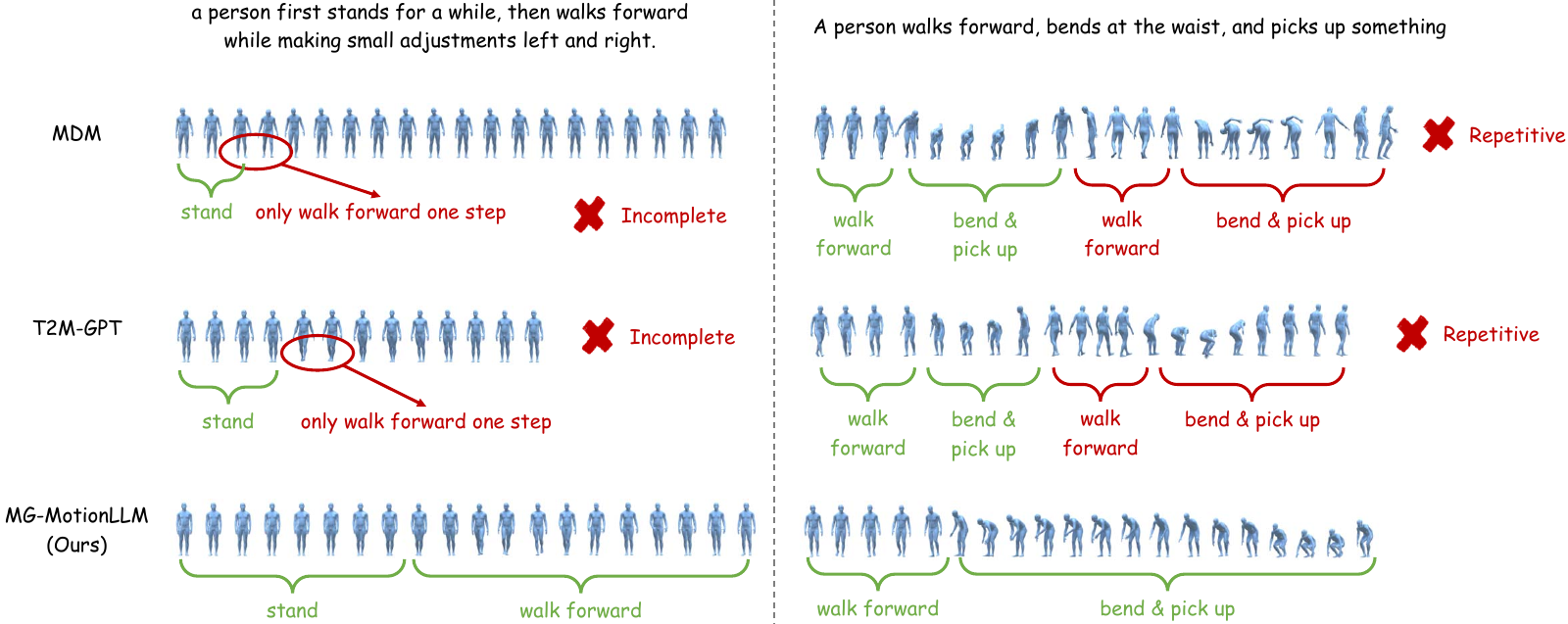}
\end{center}
\vspace{-1em}
\caption{
    \textbf{Qualitative comparison of the classical methods in the text-to-motion task.} 
    Our MG-MotionLLM produces high-quality motions that strictly match the textual descriptions.
}
\label{fig:t2m_examples}
\end{figure*}

Fig.\ref{fig:t2m_examples} shows qualitative results of our MG-MotionLLM (Granularity-Synergy Pre-trained) model on the text-to-motion task. 
Compared to state-of-the-art methods specifically designed for this task, such as T2M-GPT\cite{t2mgpt} and MDM~\cite{mdm}, our model outperforms them despite being trained for only about 10,714 iterations (300,000/28). 
It can accurately follow coarse textual descriptions to generate complete, temporally ordered motions with specific frequencies.

\section{More Discussions on Text-driven Fine-grained Motion Editing}

\begin{figure*}[!b]
\begin{center}
\includegraphics[width=1.0\linewidth]{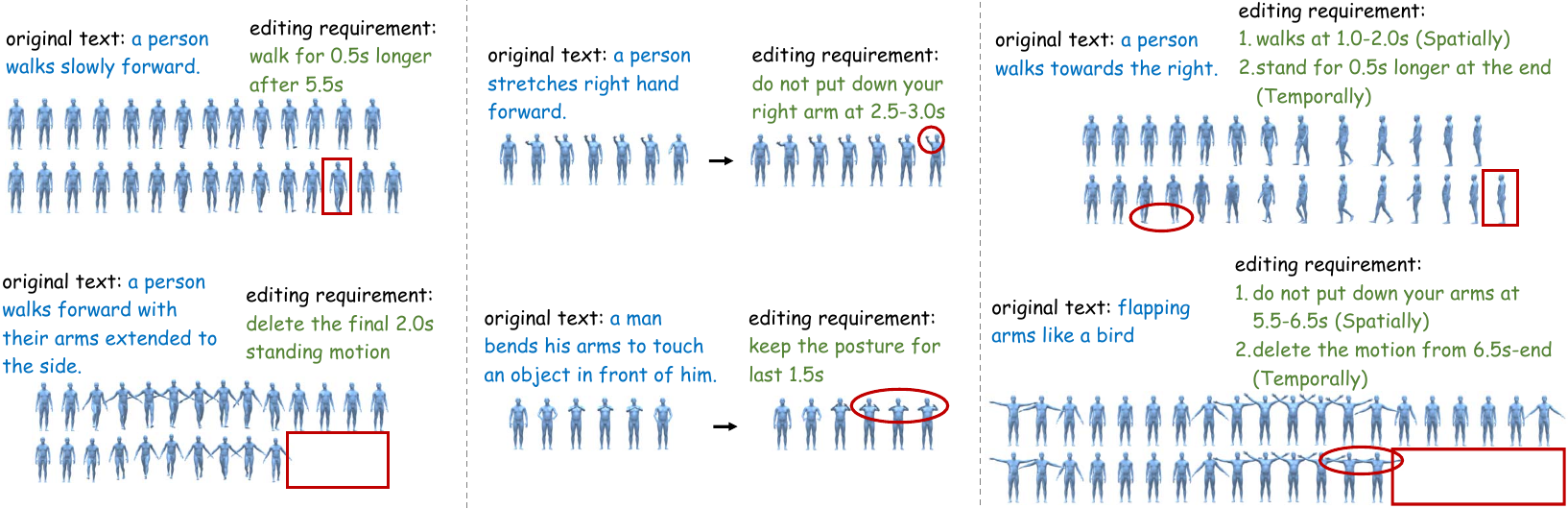}
\end{center}
\vspace{-1em}
\caption{
    \textbf{Text-Driven Fine-grained Motion Editing Examples.} 
    We display some examples of temporal editing (\textit{left}), spatial editing (\textit{middle}), and spatial-temporal editing (\textit{right}).
}
\label{fig:editing_examples_SM}
\end{figure*}

In this work, the proposed motion editing application focuses on fine-grained adjustments, such as modifying the position of an arm. 
Fig.~\ref{fig:editing_examples_SM} shows more qualitative results of text-driven fine-grained human motion editing.
In contrast, semantic changes correspond to a coarser level of granularity and can be accomplished by editing the broader motion captions.

Currently, the average motion length is 7.1 seconds, which provides approximately 14 intervals for user editing. 
We believe this level of segmentation is sufficient for users to make detailed adjustments according to their specific requirements.
At present, fine-grained motion editing requires users to manually modify the motion script. However, this process could be automated by leveraging LLMs to intelligently map user instructions to the corresponding script modifications.


\onecolumn
\section{Complete Task List in Granularity-Synergy Pre-training Stage}
\label{sec:more_tasks}

In the Granularity-Synergy Pre-training stage, we pretrain the model with a total of \textbf{28} distinct motion-relevant tasks, 
including 12 existing classical coarse-grained tasks and 16 newly proposed fine-grained ones.
We summarize the information involved into three types: 
(1) textual descriptions (including both coarse captions and detailed texts), 
(2) temporal information, and 
(3) motion data.
All the tasks are divided into three groups according to the number of information types used in the input.
We display them separately in Tab.~\ref{tab:SM_tasks_one_input}, Tab.~\ref{tab:SM_tasks_two_input}, and Tab.~\ref{tab:SM_tasks_three_input}.

\footnotesize
\SetTblrStyle{caption-tag}{\small}
\DefTblrTemplate{caption-sep}{default}{.\enskip}

\begin{longtblr}
[
    caption={\small Examples of prompt templates for tasks that utilize \textbf{one} type of information in the input.},
    label={tab:SM_tasks_one_input},
]
{
    colspec={X[halign=l, valign=m] X[2, halign=l, valign=m] X[halign=l, valign=m]},
    width=\linewidth,
    rowhead=1,
    row{even} = {bg=gray9},
    row{1} = {bg=white},
    stretch=-1,
    hline{1}={1pt,solid},
    hline{2}={0.5pt,solid},
    hline{9}={1pt,solid},
}

\textbf{Task} & \textbf{Input} & \textbf{Output} \\


Text-to-Motion & 
\begin{minipage}{\linewidth} 
    \begin{itemize}
    \setlength{\itemsep}{0.3em}
        \item Show me a motion that conveys the meaning of [caption]. 
        \item Please create a motion that represents the power of [caption]. 
        \item Give me a gesture that corresponds to [caption].
    \end{itemize}
\end{minipage} 
& [motion] \\

Motion-to-Text & 
\begin{minipage}{\linewidth}
    \begin{itemize}
    \setlength{\itemsep}{0.3em}
        \item Describe the motion portrayed in [motion] using words. 
        \item Please describe the movement shown in [motion] using words. 
        \item What type of motion does [motion] depict?
    \end{itemize}
\end{minipage} 
& [caption] \\

Motion-to-(Text, Motion Script) & 
\begin{minipage}{\linewidth}
    \begin{itemize}
    \setlength{\itemsep}{0.3em}
        \item Explain the movement depicted in [motion] with the motion summary as well as the motion script. 
        \item Depict the movement in [motion] using the motion summary and the motion script. 
        \item Illustrate the action shown in [motion] using the motion summary and the motion script.
    \end{itemize}
\end{minipage} 
& 
\begin{minipage}{\linewidth}
    \#\#\# Motion Summary \#\#\# \\ \text{[caption]} \\ 
    \text{} \\  
    \#\#\# Motion Script \#\#\# \\ \text{[motion script]}
\end{minipage} 
\\

Motion-to-Motion Script &
\begin{minipage}{\linewidth}
    \begin{itemize}
    \setlength{\itemsep}{0.3em}
        \item Explain the movement depicted in [motion] with the motion script. 
        \item Illustrate the action shown in [motion] using the motion script. 
        \item What is the motion in [motion]? Describe it using the motion script.
    \end{itemize}
\end{minipage} 
& 
\begin{minipage}{\linewidth}
\#\#\# Motion Script \#\#\# \\ 
\text{[motion script]}
\end{minipage} 
\\

(Motion Script, Snippet Motion Script)-to-Time &
\begin{minipage}{\linewidth}
    \begin{itemize}
    \setlength{\itemsep}{0.6em}
        \item {Determine the start and end times of the snippet of the motion script within the whole motion script. \\ \#\#\# Whole Motion Script \#\#\# \\ \text{[motion script]} \\ \text{} \#\#\# Snippet Motion Script \#\#\# \\ \text{[snippet motion script]}}
        \item {Please outline the start and end points for the snippet of the motion script within the whole motion script. \\ \#\#\# Whole Motion Script \#\#\# \\ \text{[motion script]} \\ \text{} \#\#\# Snippet Motion Script \#\#\# \\ \text{[snippet motion script]}}
        \item {Could you detail the timing for the snippet of the motion script as it appears in the whole motion script? \\ \#\#\# Whole Motion Script \#\#\# \\ \text{[motion script]} \\ \text{} \#\#\# Snippet Motion Script \#\#\# \\ \text{[snippet motion script]}}
    \end{itemize}
\end{minipage} 
& [time] \\

(Motion, Snippet Motion)-to-Time & 
\begin{minipage}{\linewidth}
    \begin{itemize}
        \setlength{\itemsep}{0.3em}
        \item What are the time markers for [snippet motion] within [motion]?
        \item Provide the start and finish times of [snippet motion] within [motion].
        \item Outline the time span of [snippet motion] within the context of [motion].
    \end{itemize}
\end{minipage} 
& [time] \\

(Text, Motion Script)-to-Motion & 
\begin{minipage}{\linewidth}
    \begin{itemize}
    \setlength{\itemsep}{0.6em}
        \item {Please create a motion that represents the power of the motion summary and adheres to the motion script. \\ \#\#\# Motion Summary \#\#\# \\ \text{[caption]} \\ \text{} \#\#\# Motion Script \#\#\# \\ \text{[motion script]}}
        \item {Show me a motion that captures the essence of the motion summary and reflects the motion script. \\ \#\#\# Motion Summary \#\#\# \\ \text{[caption]} \\ \text{} \#\#\# Motion Script \#\#\# \\ \text{[motion script]}}
        \item {Design a motion that embodies the emotion of the motion summary and follows the motion script. \\ \#\#\# Motion Summary \#\#\# \\ \text{[caption]} \\ \text{} \#\#\# Motion Script \#\#\# \\ \text{[motion script]}}
    \end{itemize}
\end{minipage} 
& [motion] 

\end{longtblr}

\begin{longtblr}
[
    caption={\small Examples of prompt templates for tasks that utilize \textbf{two} types of information in the input.},
    label={tab:SM_tasks_two_input},
]
{
    colspec={X[halign=l, valign=m] X[2, halign=l, valign=m] X[halign=l, valign=m]},
    width=\linewidth,
    rowhead=1,
    row{even} = {bg=gray9},
    row{1} = {bg=white},
    stretch=-1,
    hline{1}={1pt,solid},
    hline{2}={0.5pt,solid},
    hline{17}={1pt,solid},
}

\textbf{Task} & \textbf{Input} & \textbf{Output} \\


(Time, Motion Script)-to-Snippet Motion Script & 
\begin{minipage}{\linewidth}
    \begin{itemize}
    \setlength{\itemsep}{0.6em}
        \item {What is [time]'s content in the whole motion script? \\ \#\#\# Whole Motion Script \#\#\# \\ \text{[motion script]}}
        \item {Detail [time] in the scope of the whole motion script. \\ \#\#\# Whole Motion Script \#\#\# \\ \text{[motion script]}}
        \item {Show the details of [time] within the whole motion script. \\ \#\#\# Whole Motion Script \#\#\# \\ \text{[motion script]}}
    \end{itemize} 
\end{minipage}
& {\#\#\# \text{[time]}'s Motion Script \#\#\# \\ \text{[snippet motion script]}}   \\

(Time, Motion)-to-Snippet Motion &
\begin{minipage}{\linewidth}
    \begin{itemize}
    \setlength{\itemsep}{0.3em}
        \item Illustrate the movement for [time] in the scope of [motion].
        \item Capture the motion for [time] within [motion].
        \item What does the movement of [time] look like in [motion]?
    \end{itemize} 
\end{minipage}
& [snippet motion] \\

(Motion, Snippet Motion Script)-to-Time &
\begin{minipage}{\linewidth}
    \begin{itemize}
    \setlength{\itemsep}{0.6em}
        \item {What are the start and end times of the snippet of the motion script in the [motion]? \\ \#\#\# Motion Script \#\#\# \\ \text{[snippet motion script]}}
        \item {Please outline the start and end points for the snippet of the motion script within the [motion]. \\ \#\#\# Motion Script \#\#\# \\ \text{[snippet motion script]}}
        \item {Could you detail the start and end times of the snippet of the motion script as found in the [motion]? \\ \#\#\# Motion Script \#\#\# \\ \text{[snippet motion script]}}
    \end{itemize} 
\end{minipage}
& [time] \\

(Motion Script, Snippet Motion)-to-Time &
\begin{minipage}{\linewidth}
    \begin{itemize}
    \setlength{\itemsep}{0.6em}
        \item {Can you pinpoint the duration of [snippet motion] within the motion guided by the motion script?  \\ \#\#\# Motion Script \#\#\# \\ \text{[motion script]}}
        \item {What are the beginning and end points of [snippet motion] in the motion following the motion script?    \\ \#\#\# Motion Script \#\#\# \\ \text{[motion script]}}
        \item {Can you indicate the start and stop times for [snippet motion] in the sequence aligned with the motion script?  \\ \#\#\# Motion Script \#\#\# \\ \text{[motion script]}}
    \end{itemize} 
\end{minipage}
& [time] \\

(Text, Head Motion)-to-Motion &
\begin{minipage}{\linewidth}
    \begin{itemize}
    \setlength{\itemsep}{0.3em}
        \item Create a gesture that starts with [head motion] and embodies [caption].  
        \item Start with [head motion] and generate a movement that captures [caption].
        \item Initiate a gesture with [head motion] that signifies [caption]. 
    \end{itemize} 
\end{minipage}
& [motion] \\

(Text, Tail Motion)-to-Motion & 
\begin{minipage}{\linewidth}
    \begin{itemize}
    \setlength{\itemsep}{0.3em}
        \item Create a motion that concludes with [tail motion] and represents [caption]. 
        \item Generate a gesture that conveys [caption] and concludes with [tail motion]. 
        \item Show a gesture that captures [caption] and finishes with [tail motion].
    \end{itemize}
\end{minipage}
& [motion] \\

(Text, Random Motions)-to-Motion &
\begin{minipage}{\linewidth}
    \begin{itemize}
    \setlength{\itemsep}{0.3em}
        \item Design a gesture using the tokens [random motions] to express [caption].  
        \item Craft a gesture reflecting [caption] through the tokens [random motions]. 
        \item Form a gesture reflecting [caption] with the tokens [random motions].
    \end{itemize}
\end{minipage}
& [motion] \\

(Text, Motion Script, Head Motion)-to-Motion &
\begin{minipage}{\linewidth}
    \begin{itemize}
    \setlength{\itemsep}{0.6em}
        \item {Starting with [head motion], create a motion that aligns with the motion summary and adheres to the motion script. \\ \#\#\# Motion Summary \#\#\# \\ \text{[caption]} \\ \text{} \#\#\# Motion Script \#\#\# \\ \text{[motion script]}}
        \item {Initiate a motion with [head motion] that matches the motion summary and follows the motion script. \\ \#\#\# Motion Summary \#\#\# \\ \text{[caption]} \\ \text{} \#\#\# Motion Script \#\#\# \\ \text{[motion script]}}
        \item {From the initial [head motion], develop a motion that complies with the motion summary and follows the motion script. \\ \#\#\# Motion Summary \#\#\# \\ \text{[caption]} \\ \text{} \#\#\# Motion Script \#\#\# \\ \text{[motion script]}}
    \end{itemize}
\end{minipage}
& [motion] \\

(Text, Motion Script, Tail Motion)-to-Motion &
\begin{minipage}{\linewidth}
    \begin{itemize}
    \setlength{\itemsep}{0.6em}
        \item {Create a motion that reflects the motion summary and adheres to the motion script, ending with [tail motion]. \\ \#\#\# Motion Summary \#\#\# \\ \text{[caption]} \\ \text{} \#\#\# Motion Script \#\#\# \\ \text{[motion script]}}
        \item {Produce a motion that embodies the motion summary and matches the motion script, concluding with [tail motion].  \\ \#\#\# Motion Summary \#\#\# \\ \text{[caption]} \\ \text{} \#\#\# Motion Script \#\#\# \\ \text{[motion script]}} 
        \item {Craft a motion that illustrates the motion summary, follows the motion script, and ends with [tail motion].  \\ \#\#\# Motion Summary \#\#\# \\ \text{[caption]} \\ \text{} \#\#\# Motion Script \#\#\# \\ \text{[motion script]}}
    \end{itemize}
\end{minipage}
& [motion] \\

(Text, Motion Script, Random Motions)-to-Motion &
\begin{minipage}{\linewidth}
    \begin{itemize}
    \setlength{\itemsep}{0.6em}
        \item {Construct a motion with the tokens [random motions] that matches the motion summary and adheres to the motion script. \\ \#\#\# Motion Summary \#\#\# \\ \text{[caption]} \\ \text{} \#\#\# Motion Script \#\#\# \\ \text{[motion script]}}
        \item {Design a movement with key tokens [random motions] that conveys the motion summary and follows the motion script.  \\ \#\#\# Motion Summary \#\#\# \\ \text{[caption]} \\ \text{} \#\#\# Motion Script \#\#\# \\ \text{[motion script]}}
        \item {Formulate a motion with the tokens [random motions] that conveys the motion summary and follows the motion script.  \\ \#\#\# Motion Summary \#\#\# \\ \text{[caption]} \\ \text{} \#\#\# Motion Script \#\#\# \\ \text{[motion script]}}
    \end{itemize}
\end{minipage}
& [motion] \\

(Text, Time)-to-Motion &
\begin{minipage}{\linewidth}
    \begin{itemize}
    \setlength{\itemsep}{0.3em}
        \item Can you generate a [time] segment of the movement that embodies [caption]?
        \item Give me [time] of the motion that reflects the meaning of [caption].  
        \item Can you produce a motion segment of [time] representing [caption]? 
    \end{itemize}
\end{minipage}
& [motion] \\

(Motion, Time)-to-Snippet Motion Script &
\begin{minipage}{\linewidth}
    \begin{itemize}
    \setlength{\itemsep}{0.3em}
        \item Detail the motion for [time] in [motion], using the motion script.
        \item Describe the movement for [time] in [motion], with the motion script. 
        \item Clarify the action for [time] in [motion] using the motion script.
    \end{itemize}
\end{minipage}
& { \#\#\# Motion Script \#\#\# \\ \text{[snippet motion script]}} \\

(Text, Snippet Motion)-to-Time &
\begin{minipage}{\linewidth}
    \begin{itemize}
    \setlength{\itemsep}{0.3em}
        \item Pinpoint the exact times for [snippet motion] within the motion sequence that captures [caption]. 
        \item Identify when the segment [snippet motion] starts and finishes in the motion expressing [caption]. 
        \item Can you identify the timing of the snippet [snippet motion] in the motion that symbolizes[caption]?
    \end{itemize}
\end{minipage}
& [time] \\

(Text, Motion Script, Snippet Motion)-to-Time &
\begin{minipage}{\linewidth}
    \begin{itemize}
    \setlength{\itemsep}{0.6em}
         \item {Identify the timing of the snippet [snippet motion] within the motion that reflects the motion summary and adheres to the motion script.   \\ \#\#\# Motion Summary \#\#\# \\ \text{[caption]} \\ \text{} \#\#\# Motion Script \#\#\# \\ \text{[motion script]}}
         \item {Detail the start and end times for the segment [snippet motion] in the motion that matches the motion summary and follows the motion script.  \\ \#\#\# Motion Summary \#\#\# \\ \text{[caption]} \\ \text{} \#\#\# Motion Script \#\#\# \\ \text{[motion script]}}
         \item {Could you detail the timing of the segment [snippet motion] in the gesture that complies with the motion summary and follows the motion script?  \\ \#\#\# Motion Summary \#\#\# \\ \text{[caption]} \\ \text{} \#\#\# Motion Script \#\#\# \\ \text{[motion script]}}
    \end{itemize}
\end{minipage}
& [time] \\

(Text, Motion Script, Time)-to-Motion &
\begin{minipage}{\linewidth}
    \begin{itemize}
    \setlength{\itemsep}{0.6em}
        \item {Can you generate a [time] segment of the movement that embodies the motion summary and the motion script? \\ \#\#\# Motion Summary \#\#\# \\ \text{[caption]} \\ \#\#\# Motion Script \#\#\# \\ \text{[motion script]}}
        
        \item {Give me [time] of the motion that reflects the meaning of the motion summary and the motion script.  \\ \#\#\# Motion Summary \#\#\# \\ \text{[caption]} \\ \#\#\# Motion Script \#\#\# \\ \text{[motion script]}}
        
        \item {Given the motion summary and the motion script, can you give me its [time] clip?  \\ \#\#\# Motion Summary \#\#\# \\ \text{[caption]} \\ \#\#\# Motion Script \#\#\# \\ \text{[motion script]}}
    \end{itemize}
\end{minipage}
& [motion]

\end{longtblr}

\begin{longtblr}
[
    caption={\small Examples of prompt templates for tasks that utilize \textbf{three} types of information in the input.},
    label={tab:SM_tasks_three_input},
]
{
    colspec={X[halign=l, valign=m] X[2, halign=l, valign=m] X[halign=l, valign=m]},
    width=\linewidth,
    rowhead=1,
    row{even} = {bg=gray9},
    row{1} = {bg=white},
    stretch=-1,
    hline{1}={1pt,solid},
    hline{2}={0.5pt,solid},
    hline{8}={1pt,solid},
}

\textbf{Task} & \textbf{Input} & \textbf{Output} \\


(Text, Time, Head Motion)-to-Motion &
\begin{minipage}{\linewidth}
    \begin{itemize}
    \setlength{\itemsep}{0.3em}
        \item For a gesture that aligns with [caption], provide [time]'s motion snippet that begins with [head motion].
        \item Please produce a [time] motion segment, using [head motion] as the start point, from a gesture representing [caption]. 
        \item I need a [time] snippet with [head motion] as the initial input and from a gesture that symbolizes [caption].
    \end{itemize}
\end{minipage}
& [motion] \\

(Text, Time, Tail Motion)-to-Motion &
\begin{minipage}{\linewidth}
    \begin{itemize}
    \setlength{\itemsep}{0.3em}
        \item Create a [time] motion clip ending with [tail motion] from a gesture reflecting [caption].
        \item Generate a [time] motion clip that ends with [tail motion] from a gesture symbolizing [caption]. 
        \item Produce a [time] motion segment ending with [tail motion] from a gesture that represents [caption].
    \end{itemize}
\end{minipage}
& [motion] \\

(Text, Time, Random Motions)-to-Motion &
\begin{minipage}{\linewidth}
    \begin{itemize}
    \setlength{\itemsep}{0.3em}
        \item Create a [time] snippet featuring [random motions] from a motion that signifies [caption].
        \item Deliver a [time] excerpt including [random motions] from a gesture reflecting [caption]. 
        \item Create a [time] snippet with key tokens [random motions] from a gesture reflecting [caption]. 
    \end{itemize}
\end{minipage}
& [motion] \\

(Text, Motion Script, Time, Head Motion)-to-Motion &
\begin{minipage}{\linewidth}
    \begin{itemize}
    \setlength{\itemsep}{0.6em}
        \item {I need a [time]\ snippet, starting at [head motion]\, from the motion detailed in the motion summary and based on the motion summary.  \\ \#\#\# Motion Summary \#\#\# \\ \text{[caption]} \\ \#\#\# Motion Script \#\#\# \\ \text{[motion script]}}
        
        \item {Produce a [time]\ clip, starting with [head motion]\, taken from the motion defined by the motion summary and the motion summary.  \\ \#\#\# Motion Summary \#\#\# \\ \text{[caption]} \\ \#\#\# Motion Script \#\#\# \\ \text{[motion script]}}
        
        \item {Generate a [time]\ motion clip beginning with [head motion], sourced from the motion outlined in the motion summary and structured by the motion summary.  \\ \#\#\# Motion Summary \#\#\# \\ \text{[caption]} \\ \#\#\# Motion Script \#\#\# \\ \text{[motion script]}}
    \end{itemize}
\end{minipage}
& [motion] \\

(Text, Motion Script, Time, Tail Motion)-to-Motion &
\begin{minipage}{\linewidth}
    \begin{itemize}
    \setlength{\itemsep}{0.6em}
        \item {Provide a [time] clip that ends with [tail motion], derived from the full gesture that represents the motion summary and follows the motion script.  \\ \#\#\# Motion Summary \#\#\# \\ \text{[caption]} \\ \#\#\# Motion Script \#\#\# \\ \text{[motion script]}}
        
        \item {Produce a [time] snippet, concluding with [tail motion], from the motion that mirrors the motion summary and is built according to the motion script.  \\ \#\#\# Motion Summary \#\#\# \\ \text{[caption]} \\ \#\#\# Motion Script \#\#\# \\ \text{[motion script]}}
        
        \item {Please provide a [time] snippet that ends with [tail motion], from the entire motion described by the motion summary and follows the motion script.  \\ \#\#\# Motion Summary \#\#\# \\ \text{[caption]} \\ \#\#\# Motion Script \#\#\# \\ \text{[motion script]}}
    \end{itemize}
\end{minipage}
& [motion] \\

(Text, Motion Script, Time, Random Motions)-to-Motion &
\begin{minipage}{\linewidth}
    \begin{itemize}
    \setlength{\itemsep}{0.6em}
        \item {Create a [time] motion snippet with [random motions], based on a gesture that matches the motion summary and adheres to the motion script.  \\ \#\#\# Motion Summary \#\#\# \\ \text{[caption]} \\ \#\#\# Motion Script \#\#\# \\ \text{[motion script]}}
        
        \item {Create a [time] motion snippet featuring [random motions], based on a gesture that matches the motion summary and follows the motion script.   \\ \#\#\# Motion Summary \#\#\# \\ \text{[caption]} \\ \#\#\# Motion Script \#\#\# \\ \text{[motion script]}}
        
        \item {Generate a [time] motion clip with [random motions], originating from a gesture that matches the motion summary and follows the motion script.  \\ \#\#\# Motion Summary \#\#\# \\ \text{[caption]} \\ \#\#\# Motion Script \#\#\# \\ \text{[motion script]}}
    \end{itemize}
\end{minipage}
& [motion]

\end{longtblr}

\normalsize

\vspace{1em}
\begin{multicols}{2}

\section{Limitations and Future Work}

To the best of our knowledge, this work is the first to explore human motion comprehension and generation with language across multiple levels of granularity. 
However, the proposed MG-MotionLLM has certain limitations, which point to promising directions for future research.

First, MG-MotionLLM primarily focuses on the body movements of human subjects, leaving finer details such as facial expressions and hand gestures unexplored. 
Extending the model to capture these aspects could offer a more comprehensive and holistic understanding of human motion.

Second, MG-MotionLLM could be extended to encompass diverse scenarios, including multi-human interactions, animal motions, and human-object interactions, thereby broadening its applicability.

Third, the current fine-grained motion editing requires users to manually edit the motion script. Future work could explore reducing manual intervention by leveraging LLMs to intelligently bridge concise user instructions with corresponding motion script modifications, making the editing process more intuitive and efficient.

Finally, in addition to textual descriptions at finer granularities, future research could integrate more nuanced control modalities, such as short musical compositions. 
This enhancement would significantly expand the model's potential for real-world applications.

\end{multicols}

\end{appendices}

\end{document}